\documentclass[12pt, a4paper]{article}

\usepackage[numbers]{natbib}
\usepackage{stfloats}
\usepackage{mathrsfs}
\usepackage{amsmath}
\usepackage{color}
\usepackage[dvips]{graphicx}
\usepackage{subfigure}
\usepackage{multirow}
\usepackage{setspace}
\usepackage[left]{lineno}
\usepackage{tabularx}
\newcommand{\MVC}{\textrm{MVC}}
\newcommand{\joint}{\textrm{joint}}
\newcommand{\cem}{\textrm{cem}}
\newcommand{\shoulder}{\textrm{shoulder}}
\newcommand{\elbow}{\textrm{elbow}}
\newcommand{\push}{\textrm{push}}
\newcommand{\pull}{\textrm{pull}}
\newcommand{\load}{\textrm{load}}



\begin{document}%

\setcounter{page}{1}










\title{Predictive model of the human muscle fatigue: application to repetitive push-pull tasks with light external load}

\author{Sophie Sakka$^1$, Damien Chablat$^2$, Ruina Ma$^3$ and Fouad Bennis$^3$ \\
IRCCyN and University of Poitiers, France\\
E-mail: Sophie.Sakka@irccyn.ec-nantes.fr \\
IRCCyN and CNRS, France \\
E-mail: Damien.Chablat@irccyn.ec-nantes.fr \\
IRCCyN and \'Ecole Centrale de Nantes, France\\
E-mail: Fouad.Bennis@ec-nantes.fr}

\maketitle
\begin{abstract}
Repetitive tasks in industrial works may contribute to health problems among operators, such as musculo-skeletal disorders, in part due to insufficient control of muscle fatigue. In this paper, a predictive model of fatigue is proposed for repetitive push/pull operations. Assumptions generally accepted in the literature are first explicitly set in this framework. Then, an earlier static fatigue model is recalled and extended to quasi-static situations. Specifically, the maximal torque that can be generated at a joint is not considered as constant, but instead varies over time accordingly to the operator's changing posture. The fatigue model is implemented with this new consideration and evaluated in a simulation of push/pull operation.
{\bf Keywords}:\it{Predictive model; Fatigue; Push/pull operations; Repetitive tasks; simulation}
\end{abstract}
\section{Introduction}
A better understanding on how the human body deals with muscle fatigue would allow better estimates of working abilities and enhance production. Such understanding could also help in occupational health management, such as reducing work-related musculo-skeletal disorders (MSDs)~\cite{hoozemans1998,niosh1981,snook1978,klein1984}. While the role of muscle fatigue in MSDs is not well understood, it has been established that when muscles fatigue the torque production strategy changes~\cite{White1990}. For instance, the body has to use muscle fibers that are more vulnerable to fatigue or use different muscle groups that are less efficient to maintain the level of torque demanded. If a fatiguing operation goes on without sufficient rest, such adaptations may increase the risk of MSDs. 

Muscle fatigue should, ideally, be predicted to aid in designing or evaluating repetitive industrial operations. Such a predictive model would allow optimizing working and resting phases, and could enhance long term health. Yet, such fatigue prediction models are difficult to develop. It is well known that muscle fatigue is a complex phenomenon with multiple sources, of which the relative contributions remain unclear. For example, fatigue origins may be mechanical (peripheral fatigue: muscles are not able to respond to the input signal), neurological (central fatigue: the signal sent to the muscles is noisy or reduced), psychological, and/or other sources. As a consequence, predictive muscle fatigue models remain limited to restrictive assumptions and situations and focus on limited parts of the body.

Several models have been previously proposed for muscle fatigue prediction~\cite{Liu2002,Ma2009,Xia2008}. Most models, though, have focused on static situations: a desired torque at a fixed body configuration, which is an oversimplification of workplace tasks. In reality, the ability of torque production varies with body configuration due to the length-tension relationship. This paper proposes a more general muscle fatigue model, extending the static case to quasi-static situations, in which the subject's posture can be changing, hence a varying torque capability. It is worth noting that this new modeling approach cannot yet be considered as dynamic as the torque capability is assumed to depend primarily on the joint configuration alone. In dynamic situations, this capability depends on both joint configuration and joint motion (i.e. the force-velocity relationship in muscle force production). Such data will be required to establish a fully dynamic fatigue model in the future. 


To develop and analyze a quasi-static muscle fatigue model, we simulate arm motion during a repetitive flexion-extension task. Even though this task is common in industrial applications~\cite{travailsafe2010} and daily life activities~\cite{Kumar2008}, few studies have examined it with a varying body configuration~\cite{Mital1995,Badi2008,Bonato2002}. In this paper we first set the framework and assumptions for modeling. The arm configuration data is then simulated and used to set up the model. Finally, the implication of the study findings is discussed.  
\section{Problem setting for push/pull operations}
\subsection{Definition of Push/Pull operations}
A push/pull operation generally defines both an action of pushing during which the force generated by the hand is oriented away from the body and an action of pulling during which the force is oriented toward the body. In daily life and in industrial environments, push/pull operations are very common, such as in objects displacement including up and down lifting or left and right shifting, drilling operations, buttons pushing, and so on. These actions of pushing and/or pulling are an important source of MSD~\cite{hoozemans1998,White1990}. Indeed, NIOSH reported that about 20~\% of overuse injuries were associated to push/pull operations~\cite{niosh1981}. Moreover, almost 8~\% of back pain and 9~\% of sprains and strains at the back are associated with push/pull operations~\cite{snook1978,klein1984}. In addition, heavy loads acting on the lumbar discs were detected when performing these activities~\cite{White1990}, which could also contribute to MSDs. The International Standard ISO 11228-2~\cite{desbrosses2012} recently proposed a first initiative to limit the professional required forces in push/pull operations with the aim of reducing or prevent MSD associated with  this activity.

Push/pull operations may be divided into three kinds, depending on the groups of muscles in which action is required to perform the motion: up-down; left-right or back-and-forth movements. In this paper, we will focus on the back-and-forth  movements, which corresponds to the majority of studied tasks in the literature and the most used push/pull operations in daily life.

Another key component of a push/pull operation is the external load. While the whole body is involved in all push/pull operations, it is assumed here that, for the case of light external loads, the main groups of muscles are in the arm. Thus we focus on the arm movement with a light load push-pull task, while the movements of the rest of the body are neglected. 

\subsection{Working assumptions}\label{sec:hyp}
We will set some working assumptions to simplify the study and address the problem of fatigue modeling in the frame of repetitive drilling operations. The study is made from the point of view of external forces. Specifically, the local effects of each working muscle will not be addressed, and instead the effects at the joint level (shoulder and elbow) will be considered for pushing or pulling. Assumptions made include: 
\begin{itemize}
\item[\bf A1] Only light external loads are involved, and only the arm muscles are activated when pushing and pulling. Other muscles of the body will be neglected. 
\item[\bf A2] The arm movement will be limited to two joints: the shoulder and the elbow. The forearm, the wrist joint, and the hand will be considered as a single rigid body. This assumption is considered valid for negligible movements of the wrist, such as in drilling operations using back and forth movements.
\item[\bf A3] The group of muscles used for pushing will be considered inactive when the group of muscles used for pulling is active and vice versa. This assumption is equivalent to ignoring muscle co-contraction, or more precisely the use of antagonistic muscles does not contribute to their fatigue.
\end{itemize}

The push-pull task used here is composed of two distinct phases. The pushing phase is generated by two groups of muscles: one for shoulder flexion (deltoids) and one for elbow extension (triceps). The pulling phase is generated by the two groups of muscles: one for shoulder extension (sub scapula) and one for elbow flexion (biceps). The model of fatigue will be applied to each active group of muscles. 
\subsection{Cyclic motion}
\begin{figure}[!ht]
	\centering
		\scalebox{0.7}{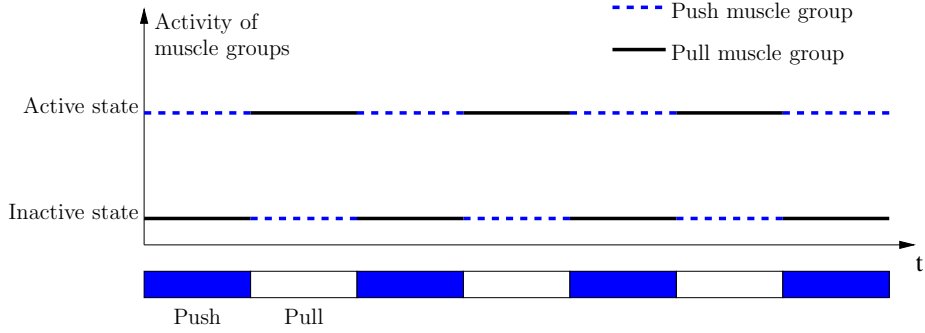}
	\caption{Representation of the group muscles activity during push/pull operations}
	\label{fig:musclegroupactivationassumption}
\end{figure}

A cyclic motion involves a repetitive movement where the beginning and ending positions, velocities and accelerations are identical. Let us consider a repetitive push/pull cycle with a regular periodicity noted $T$, such as the one illustrated in Fig.~\ref{fig:musclegroupactivationassumption}. Let $T^{\push}$ and $T^{\pull}$ denote respectively durations of the pushing and pulling phases durations with $T = T^{\push} + T^{\pull}$.

Let $\ell$ denote the number of completed cycles of the repetitive motion since the beginning of the work, and let $t^{\push}$ and $t^{\pull}$ denote the overall times counted from the beginning of the work respectively for pushing and for pulling. At the end of the work, we will have $\ell T=t_f-t_0$ where $t_0$ and $t_f$ are respectively the starting and ending instants. If the work started with a pushing phase such as illustrated in Fig.~\ref{fig:musclegroupactivationassumption}, then at the end of each pushing phase we have $t^{\push}=(\ell+1)T^{\push}$ and at the end of each pulling phase we have $t^{\pull}=\ell T^{\pull}$.

For an instant $t\in[t_0,t_f]$, we have $t=t^{\push}+t^{\pull}$. Two situations may occur, depending if $t$ is in a pushing or a pulling phase. If $t$ is in a pushing phase, then $t^{\push} = t - t^{\pull} = t - \ell\,T^{\pull}$. The corresponding function for $t^{\push}$ may be expressed using $\ell = (t_f-t_0)/T$.
\begin{equation}\label{eq:tpush}
	t^{\push} = t - \left [ \frac{t_f-t_0}{T}\right ] \; T^{\pull} 
\end{equation}
If $t$ is in a pulling phase, then we have $t^{\pull} = t - t^{\push}= t - (\ell+1) \cdot T^{\push}$, which may be expressed with the following form. 
\begin{equation}\label{eq:tpull}
	t^{\pull} = t - \left [ \frac{t_f-t_0}{T}+1\right ] \; T^{\push} 
\end{equation}
\section{Muscle fatigue model for push/pull operations}
\subsection{Formulation of general muscle fatigue behavior}
Using the assumptions in section~\ref{sec:hyp}, each group of muscles and its fatigue may be studied and its fatigue evaluated separately from the other groups. The fatigue model proposed in a previous study \cite{Ma2009} is adopted and defined as the following differential equation.
\begin{equation}\label{eq:fatiguediff}
\frac{d\Gamma_{\cem}}{dt} = -k \cdot \frac{\Gamma_{\joint}}{\Gamma_{\MVC}} \cdot \Gamma_\cem
\end{equation}
where $k$ denotes a constant fatigue parameter and $\Gamma_{\cem}$ is the current maximal capacity for a group of muscles to generate a joint torque, and is considered as a characteristic value to evaluate the muscles fatigue. $\Gamma_{\MVC}$ ({\it Maximal Voluntary Contraction}) is the initial value of $\Gamma_{\cem}$, measured at the beginning of the experiment. Its value depends on the operator's anthropometry and body configuration as will be detailed in section~\ref{sec:gammaMVC}. 

$\Gamma_{\joint}$ denotes the desired torque vector at the considered joint. It is composed of a portion  coming from the body movements $\Gamma_{i}$ and a portion  coming from the external load $\Gamma_{i,\textrm{ext}}$ introduced as separate variables:
\begin{equation}
	\Gamma_{\joint} = \Gamma_{i} + \Gamma_{i,\textrm{ext}}\qquad\qquad i = 1,\dots,n
	\label{fig:lagrangeforceexterne}
\end{equation}
with $n$ being the number of joints observed in the motion ($n=2$ when only considering the 
s). $\Gamma_{i}$ ($i=1..n$) are calculated from the values of $\boldsymbol{\theta}$, $\dot{\boldsymbol{\theta}}$ and $\ddot{\boldsymbol{\theta}}$ using the Lagrangian formalism~\cite{Khalil2010}.
\begin{equation}
	\Gamma_{i} = \dfrac{d}{dt} \dfrac{\partial{L}}{\partial{\dot{\theta}_{i}}} - \dfrac{\partial{L}}{\partial{\theta_{i}}}\qquad\qquad i = 1,\dots,n
	\label{eq:lagrangeclassic}
\end{equation}
where $L=E-U$ is the Lagrangian, $E$ being the kinetic energy and $U$ being the potential energy of the complete system~(see~\cite{Ma2012} for the development applied to the current problem). $\Gamma_{i,\textrm{ext}}$ is directly related to the desired external loads grouped in $M_{\load}$. Then $\Gamma_{\joint} = \Gamma_{\joint}(\boldsymbol{\theta},\dot{\boldsymbol{\theta}},\ddot{\boldsymbol{\theta}},M_{\load})$.

Equation~\eqref{eq:fatiguediff} may be re-written in the following form.
\begin{equation}\label{eq:fatiguetemps}
\frac{\dot\Gamma_{\cem}(t)}{\Gamma_{\cem}(t)} = -k \cdot \frac{\Gamma_{\joint} \big( \boldsymbol{\theta}(t) , \dot{\boldsymbol{\theta}}(t), \ddot{\boldsymbol{\theta}}(t), M_{\load} \big)}{\Gamma_{\MVC} \big( \boldsymbol{\theta}(t) \big)}
\end{equation}
\begin{equation}\label{eq:fatigue}
	\Rightarrow\qquad\Gamma_{\cem}(t) = C\ \cdot\ e^{ - k \int_{t_0}^{t} \frac{\Gamma_{\joint}(\boldsymbol{\theta}(u) , \dot{\boldsymbol{\theta}}(u), \ddot{\boldsymbol{\theta}}(u), M_{\load})}{{\Gamma_{\MVC}(\boldsymbol{\theta}(u))}} du}
\end{equation}
where $C$ is an integration constant.
\subsection{Formulation of joint capacities}\label{sec:gammaMVC}
One of the key parameters of the fatigue model~\eqref{eq:fatigue} is the maximal torque $\Gamma_{\MVC}$. In static situations, in which a force is applied by the operator onto the environment with no body movement, the value of $\Gamma_{\MVC}$ varies according to its current body configuration. A predictive model to calculate $\Gamma_{\MVC}$ was proposed by Chaffin {\it et al.}~\cite{chaffin1999} and is summarized in Table~\ref{tab:strengthmodel} which gives the predictive functions for $\Gamma_{\MVC}$ at the elbow (index $e$) and shoulder (index $s$) joints. Two parameters are considered as inputs: the joint angle value $\theta$ and an adjustment gain $G$ that accounts for the gender of the operator. Several observations may be made based on the functions in Table~\ref{tab:strengthmodel}: 
\begin{enumerate}
\item Joint capacity behaves differently for flexion and for extension movements;
\item Predictive models for flexion are more complex than for extension;
\item Only the shoulder extension model does not depend on the configuration of another joint. 
\end{enumerate}
\begin{table}[h!]
\begin{center}
	\begin{tabular}{|l|c|c|c|}
		\hline
		\multirow{2}{*}{Joint/movement}  & \multirow{2}{*}{Joint capacity ($\Gamma_{\MVC}$)} & \multicolumn{2}{c|}{$G$}\\
		\cline{3-4}
		                							 &				   																										& Male\ \ \       & Female  \\
		\hline
		Elbow flexion & $\big( 336.29 + 1.544\theta_{e} - 0.0085\theta_{e}^{2} -0.5\theta_{s} \big) G$	& 0.1913         & 0.1005   \\
		\hline
		Elbow extension	&	$\big( 264.153 + 0.575\theta_{e} - 0.425\theta_{s} \big) G$  		& 0.2126         & 0.1153   \\	 
		\hline
		Shoulder flexion &	$\big( 227.338 + 0.525\theta_{e} - 0.296\theta_{s} \big) G$  		& 0.2854         & 0.1495   \\
		\hline
		Shoulder extension &	$\big (204.562 + 0.099\theta_{s} \big) G$     											& 0.4957         & 0.2485   \\
		\hline
	\end{tabular}
	\caption{Chaffin's model for joint capacity in static situations~\cite{chaffin1999}}
	\label{tab:strengthmodel}
\end{center}
\end{table}

The model inputs are the measured shoulder and elbow constant joint angles, with the value 0 matching a straight arm in a classical standing configuration. From the joint angles, the respective shoulder and elbow respective capacities can be estimated, assuming this motion is relatively slow. The obtained joint capacity values will then be used for the fatigue model~\eqref{eq:fatigue}.

It is worth noting that static strength data are being used here to model a dynamic task. While some dynamic strength data have been reported previously, such as in Frey {\it et al.}~\cite{Frey2012}, comprehensive dynamic strength data for the shoulder are not available. For simplicity, we used the same approach for both the elbow and the shoulder joints, based on Chaffin's predictive functions. As a consequence, the proposed model is considered ``quasi-static''.
\subsection{Formulation  of the push/pull fatigue in static situations}
The model~\eqref{eq:fatigue} is applied separately for each muscles group, for pushing and pulling. 
\begin{equation}\label{eq:musclefatigue}
	\begin{array}{lcl}
	\Gamma_{\cem}^{\push}(t^{\push}) & = & \Gamma_{\MVC}^{\push} \cdot e^{ - \frac{k}{\Gamma_{\MVC}^{\push}} \int_{0}^{t^{\push}} \Gamma_{\joint}^{\push} du}\\
	\Gamma_{\cem}^{\pull}(t^{\pull}) & = & \Gamma_{\MVC}^{\pull} \cdot e^{ - \frac{k}{\Gamma_{\MVC}^{\pull}} \int_{0}^{t^{\pull}} \Gamma_{\joint}^{\pull} du}
	\end{array}
\end{equation}

\begin{figure}[htbp]
	\centering
	\scalebox{0.7}{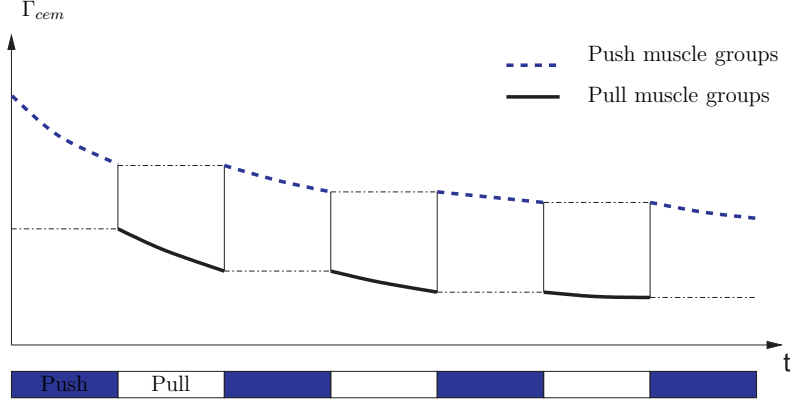}
	\caption{\label{fig:musclegroupfatigue} Schematic representation of fatigue in pushing and pulling groups of muscles ($\Gamma_{\cem}$) at a joint during a repetitive push/pull operation in quasi-static situations}
\end{figure}

Because of Assumption A3, the model of fatigue of the arm for a push/pull operation is represented by piecewise continuous functions, in which each muscle group has increasing fatigue when working and remains at the same level of fatigue when relaxing (Fig.~\ref{fig:musclegroupfatigue}). Recovery is not taken into account in this initial model development, which means that the subject should be less tired than predicted by the proposed model. Still, this model of fatigue leads to major observations on how human beings deal with fatigue independently from recovery. Depending on whether the current time $t$ is in a pushing or a pulling phase, the model takes the two cases into account as follows.
\begin{equation}
	\Gamma_{\cem}(t) =
	\begin{cases}	
		\quad\Gamma_{\cem}^{\push} (t^{\push}), & t \in \textrm{pushing phase}\\
		\quad\Gamma_{\cem}^{\pull} (t^{\pull}), & t \in \textrm{pulling phase}
	\end{cases}
	\label{eq:pushpullfatiguefirststep}
\end{equation}
When substituting \eqref{eq:tpush} and \eqref{eq:tpull} into \eqref{eq:pushpullfatiguefirststep}, the following model is obtained.
\begin{equation}
	\Gamma_{\cem}(t) =
	\begin{cases}	
		\quad\Gamma_{\cem}^{\push} \left(t - \left [ \frac{t_f-t_0}{T}\right ] \; T^{\pull}\right), & t \in \textrm{pushing phase}\\
		\quad\Gamma_{\cem}^{\pull} \left(t - \left [ \frac{t_f-t_0}{T}+1\right ] \; T^{\push}\right), & t \in \textrm{pulling phase}
	\end{cases}
	\label{eq:pushpull}
\end{equation}

It is important to note here that the two functions $\Gamma_{\cem}^{\push}$ and $\Gamma_{\cem}^{\pull}$ are continuous. On the contrary, $\Gamma_{\cem}(t)$ is not a continuous function as illustrated in Fig.~\ref{fig:musclegroupfatigue}, which illustrates the model of fatigue \eqref{eq:pushpull} and the previous comments for one joint animated by one pushing and one pulling groups of muscles during a repetitive push/pull task. Starting with the pushing phase, only the group of muscles dedicated to pushing is activated, while the pulling one is at rest (constant value of fatigue characterized by $\Gamma_{\cem}$). When in the pulling phase, only the group of muscles dedicated to pulling is at work, and the pushing one is at rest. 
\subsection{Formulation of push/pull fatigue in quasi-static situations}
Quasi-static situations denote the case when the arm is in motion and  $\Gamma_{\MVC}$ varies with joint angles as defined in the section 3.2. These values do not remain constant but evolve with time together with the evolution of the arm joint angles: $\Gamma_{\MVC}=\Gamma_{\MVC}(\boldsymbol{\theta}(t))$. $\Gamma_{\MVC}$ cannot be taken out of the integral in \eqref{eq:fatigue}. In this case, the integration constant $C$ is set by considering $\Gamma_{\cem}(0) = \Gamma_{\MVC}(\boldsymbol{\theta}_0)$ where $\boldsymbol{\theta}_0= \boldsymbol{\theta}(t_0)$. The fatigue model in quasi-static situations is then expressed as follows.
\begin{equation}\label{eq:dynamicfatigue}
	\Gamma_{\cem}(t) = \Gamma_{\MVC} (\boldsymbol{\theta}_0) \cdot e^{ - k \int_{0}^{t} \frac{\Gamma_{\joint}(\boldsymbol{\theta}(u) , \dot{\boldsymbol{\theta}}(u), \ddot{\boldsymbol{\theta}}(u), M_{\load})}{{\Gamma_{\MVC}(\boldsymbol{\theta}(u))}} du}
\end{equation}
Similarly to the previous static case, the quasi-static fatigue model \eqref{eq:dynamicfatigue} is applied successively to the pushing and pulling groups of muscles during the push/pull operation. Recovery during the resting phase is not considered. 
\begin{equation}
	\begin{array}{lcl}
	\Gamma_{\cem}^{\push}(t^{\push}) & = & \Gamma_{\MVC}^{\push}(\boldsymbol{\theta}_0) \cdot e^{ - k \int_{0}^{t^{\push}} \frac{\Gamma_{\joint}^{\push}(\boldsymbol{\theta}(u) , \dot{\boldsymbol{\theta}}(u), \ddot{\boldsymbol{\theta}}(u), M_{\load})}{\Gamma_{\MVC}^{\push}(\boldsymbol{\theta}(u)) } du}\\
	\Gamma_{\cem}^{\pull}(t^{\pull}) & = & \Gamma_{\MVC}^{\pull}(\boldsymbol{\theta}_0) \cdot e^{ - k \int_{0}^{t^{\pull}} \frac{\Gamma_{\joint}^{\pull}(\boldsymbol{\theta}(u) , \dot{\boldsymbol{\theta}}(u), \ddot{\boldsymbol{\theta}}(u), M_{\load})}{\Gamma_{\MVC}^{\pull}(\boldsymbol{\theta}(u)) } du}
		\end{array}
		\label{eq:push_and_pull_model}
\end{equation}

Then the fatigue model for push/pull operations is obtained in quasi-static situations, similarly to static situations.
\begin{equation*}
	\Gamma_{\cem}(t) =
	\begin{cases}	
		\quad\Gamma_{\cem}^{\push} \left(t - \left [ \frac{t_f-t_0}{T}\right ] \; T^{\pull}\right), & t \in \textrm{pushing phase}\\
		\quad\Gamma_{\cem}^{\pull} \left(t - \left [ \frac{t_f-t_0}{T}+1\right ] \; T^{\push}\right), & t \in \textrm{pulling phase}
	\end{cases}
\end{equation*}

\section{Application to push/pull tasks with back and forth arm movements}\label{sec:poussertireravantarrière}
\subsection{Description of the push/pull task}
The tasks were chosen according to Ma's work~\cite{Ma2009a}, in order to compare the proposed quasi-static model to the static one. The arm of a male operator of stature 188~[cm] and body mass 90~kg was modeled to simulate the performance of a repetitive push/pull task. The model of the task included the use of a tool positioned at the extremity of the arm and with mass of 2~[kg]. The initial position of the hand was defined in the sagittal plane by $P_0 = [ P_{x,0}, P_{z,0} ]$~[m] and the final position by $P_f = [ P_{x,f}, P_{z,f} ]$~[m]. The position $\{ 0, 0 \}$ corresponds to the fixed shoulder position (the origin of the reference frame). The operator generates a 20~[N] pushing force and a 10~[N] pulling force while tracking an horizontal line, back and forth. 

\begin{figure}[!ht]
	\centering
	\subfigure[\label{fig:PushPullPosture}The hand has a 20~cm amplitude]{\includegraphics[width=0.47\textwidth]{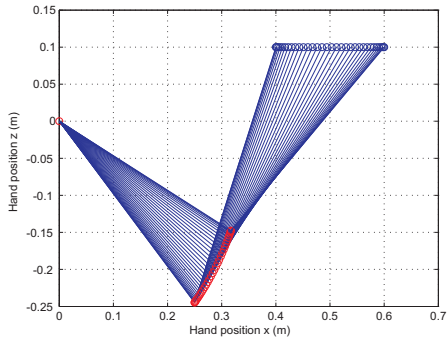}}\quad
	\subfigure[\label{fig:PushPullPostureDiscussion}The hand has a 10~cm amplitude]{\includegraphics[width=0.47\textwidth]{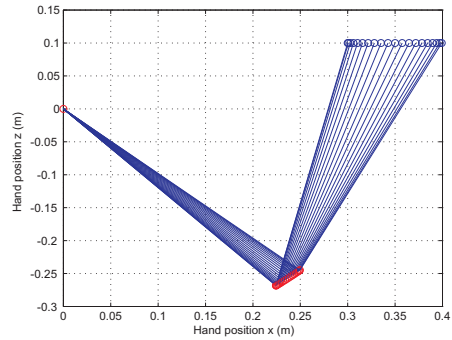}}	
	\caption{Arm movement during a drilling operation. $T_{\push} = T_{\pull} = 5$~[s], $M_{\load}^{\push}=20$~[N], $M_{\load}^{\pull}=10$~[N]}
	\label{fig:pushpulltaskperformed}
\end{figure}

The push/pull cycle lasted 10~[s], with an equivalent time distribution for pushing and for pulling: $T_{\push} = T_{\pull} = 5$~[s]. This kind of task represents an operation of classical horizontal drilling. It is illustrated in Fig.~\ref{fig:pushpulltaskperformed} for two different amplitudes of hand horizontal displacements: either with an amplitude of 20~[cm] (Fig.~\ref{fig:PushPullPosture}), or an amplitude of 10~[cm] (Fig.~\ref{fig:PushPullPostureDiscussion}). In this figure, the left red circle at $(0,0)$ represents the fixed shoulder joint; the two segments, arm and forearm, are represented by the blue solid lines linked at the elbow joint represented by a moving red circle. The hand is at the end of the kinematic chain, represented by blue circles. 
The trajectories of the elbow are different for the two tasks. 
We can see in Fig.~\ref{fig:PushPullPosture} that a 0.4 to 0.6~[m] horizontal displacement of the hand generates an elbow elevation of 10~[cm] and advancement of 7~[cm], while the 0.3 to 0.4~[m] horizontal displacement of the hand illustrated in Fig.~\ref{fig:PushPullPostureDiscussion} generates an elbow elevation of 3~[cm] and advancement of 3~[cm].
\subsection{Modeling of the operator's arm}
\begin{figure}[htbp]
	\centering
	\scalebox{0.4}{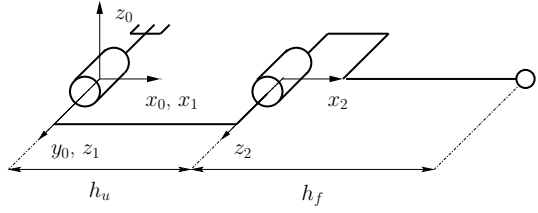}
	\caption{Kinematic representation of the operator's arm}
	\label{fig:GeometriModel}
\end{figure}

The geometric model of the arm is summarized by the modified Denavit-Hartenberg parameters~\cite{denavit1955} given in Table~\ref{tab:GeometriModel}, matching the kinematics representation in Fig.~\ref{fig:GeometriModel}. The motion only occurs in the sagittal plane $(\mathbf x_0,\mathbf z_0)$, so the shoulder and elbow joints are represented by simple rotational joints and all the rotations angles are calculated along the $\mathbf y_0$ lateral axis.

\begin{table}[!ht]
	\centering
	\begin{tabular}{|c|c|c|c|c|c|}
		\hline
		Joint	&  $\sigma$  &   $d$   & $\alpha$  &  $r$  &   $\theta $    \\
		\hline
		Shoulder    &   0       &   0    	 & $\pi/2$  	&  0     &  $\theta_s$      \\
		\hline
		Elbow     &   0       &   $h_u$  & 0  				&  0     &  $\theta_e$      \\
		\hline
	\end{tabular}
	\caption{DH parameters of the operator's arm}
	\label{tab:GeometriModel}
\end{table}

\subsection{Trajectory generation}
\begin{figure}[!ht]
	\centering	\subfigure[Elbow]{\includegraphics[width=0.45\textwidth]{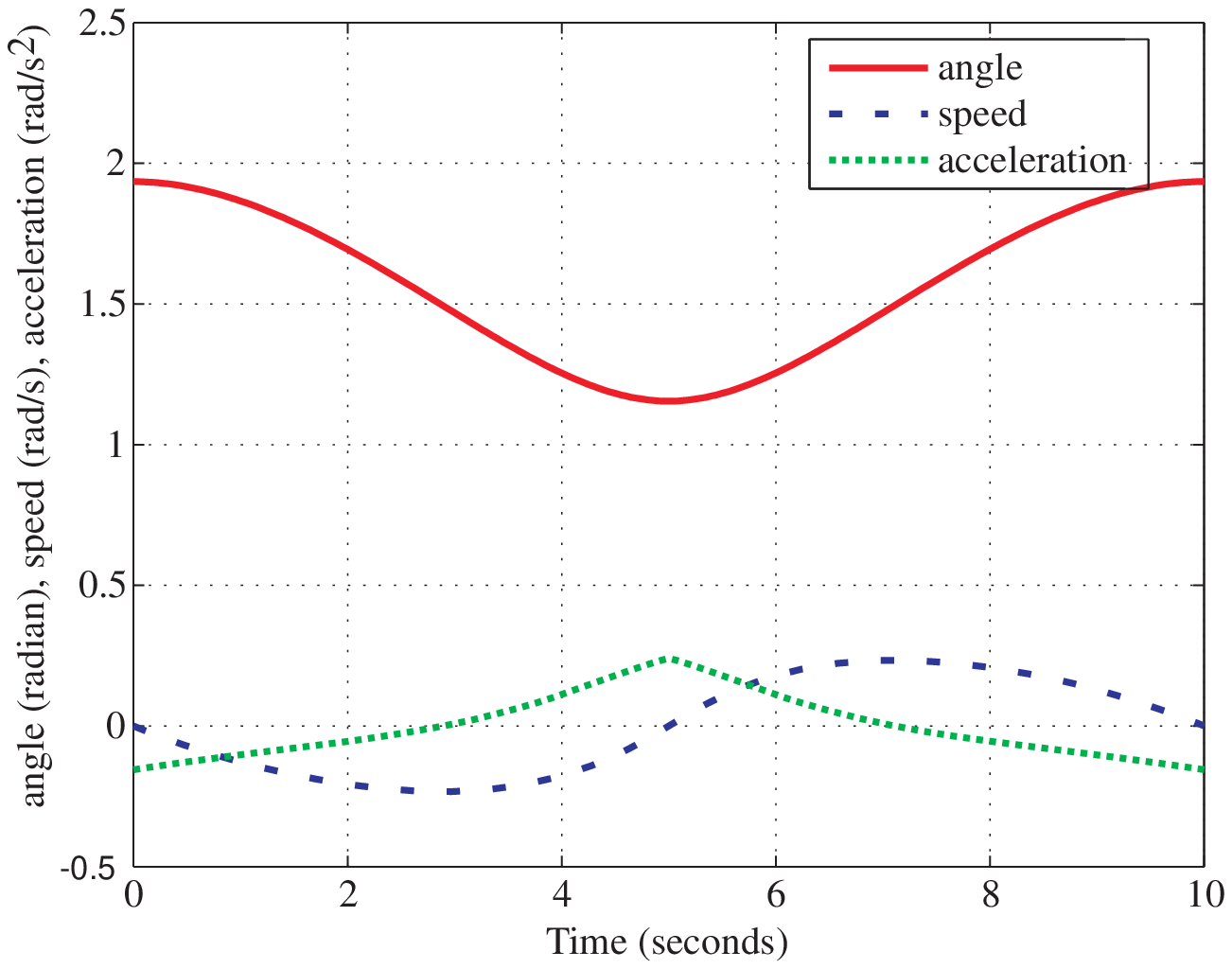}}\quad
	\subfigure[Shoulder]{\includegraphics[width=0.45\textwidth]{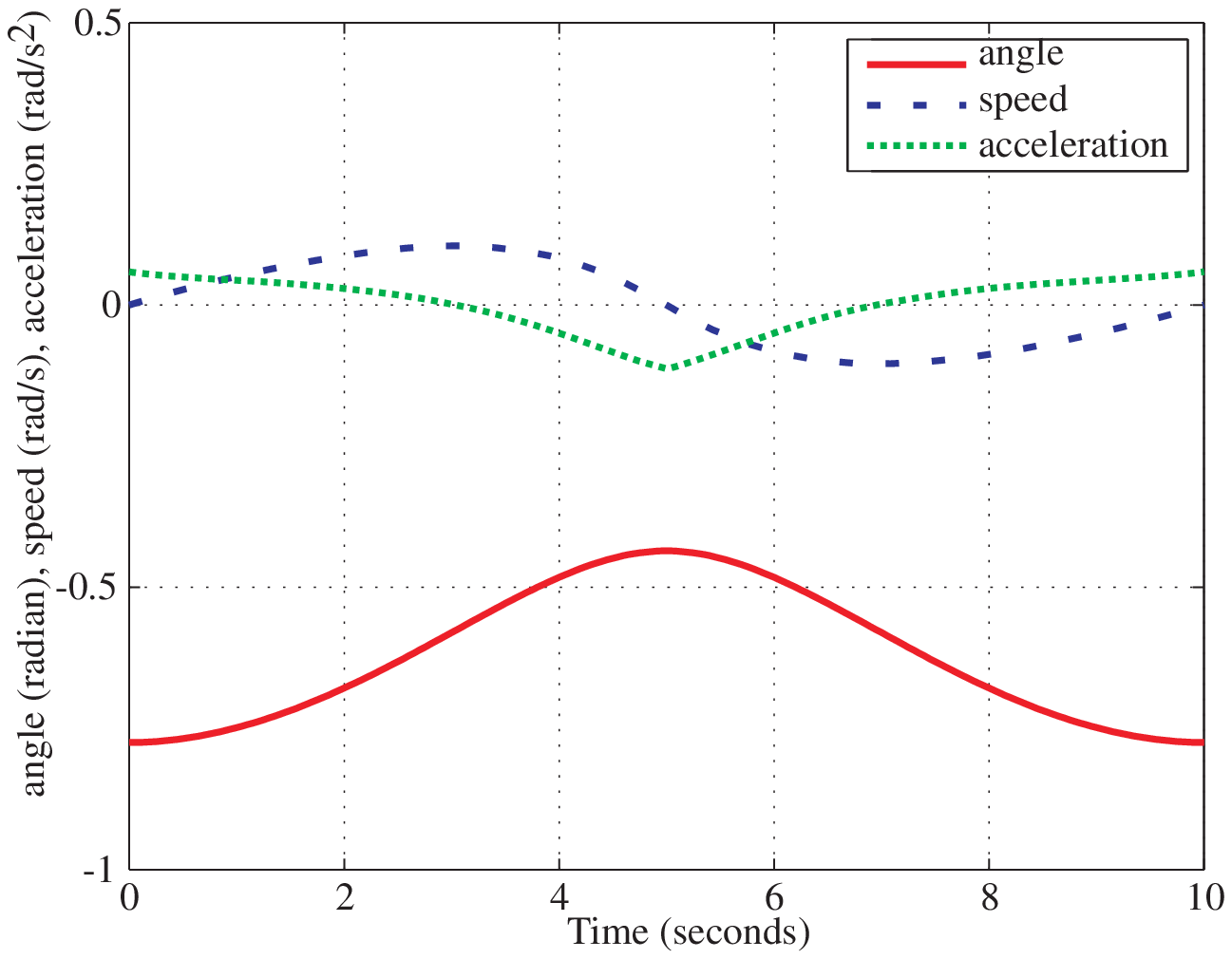}}
	\caption{Shoulder and elbow joint angles, angular velocities and angular accelerations over  time using trajectory generation in task space. The first phase is a pushing phase}
	\label{fig:trajectoryXpushpull}
\end{figure}

The hand trajectory in the task space was described using polynomial interpolation. The hand velocity and acceleration were both null at the beginning and end of each pushing and pulling phases.  Beginning and ending positions denoted respectively by  $P_0$ and $P_f$. We obtain the following trajectory.
$$ 	P(t) = P_0 + p(t) \cdot \left( P_f - P_0 \right), \quad t_0\leq t \leq t_f $$
with $	p ( t ) = 3 \left( t / t_f \right)^2 - 2 \left( t / t_f \right)^3$. The joint angles, angular velocities and angular accelerations were obtained from the hand trajectory profile using the inverse kinematics of the arm, leading to the joint profiles illustrated in Fig.~\ref{fig:trajectoryXpushpull} for one push/pull cycle. 
                            
It is worth noting that defining the kinematics in the joint space is much easier when applying the model to a real human arm, as the center of rotation of the human arm may be difficult to locate and is needed to set the inverse kinematic model of the human arm. Nevertheless, the joints profiles obtained from the definition of the kinematics in the task space offer more correspondence with the human real trajectories used to perform the described drilling operation, which means that defining the kinematics in the task space is more realistic. This kinematics description will be used in what follows. 

\begin{figure}[!ht]            
	\centering
	\includegraphics[width=0.47\textwidth]{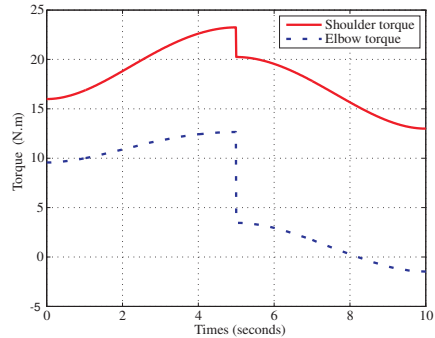}
	\caption{Shoulder and elbow joint torques $\Gamma_{\joint}$ for one operating cycle. The first phase is a pushing phase}
	\label{fig:torquedrilling}
\end{figure}

From the described kinematics, the joint torques at the shoulder, $\Gamma_{s}$, and at the elbow, $\Gamma_{e}$, can be calculated. Their respective evolution over time is illustrated in Fig.~\ref{fig:torquedrilling} for one push/pull cycle. The motion is set for an horizontal displacement of the operator's hand from $P_0=[0.4,0.1]$ to $P_f=[0.6,0.1]$ [m] which matches the case illustrated in Fig.~\ref{fig:PushPullPosture}.

Several observations may be made from Fig.~\ref{fig:torquedrilling}. First, the torque at the shoulder joint is always greater than the one at the elbow. Second, the two curves show a discontinuity at 5~[s] due to the discontinuity in external loads between the pushing (20~[N]) and pulling (10~[N]) phases. During a given phase, the torques remain continuous but not linear: their values depend on the operator's arm configuration.   
\subsection{Evolution of joint capacity $\Gamma_{\MVC}$}
\begin{figure}[!ht]            
	\centering
	\includegraphics[width=0.47\textwidth]{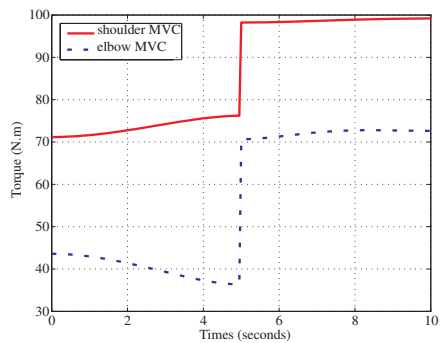}
	\caption{Evolution of $\Gamma_{\MVC}(t)$ at the shoulder and elbow joints for a push/pull cycle}
	\label{fig:drillingMVC}
\end{figure}

As mentioned in section~\ref{sec:gammaMVC}, Chaffin's work~\cite{chaffin1999} allows for determining the constant joint capacity $\Gamma_{\MVC}$ for the shoulder and the elbow according to the arm configuration and depending on the type of desired external load (pushing or pulling). We have dissociated two working phases for a complete push/pull cycle: the pushing phase, in which the shoulder is in flexion and the elbow in extension, and the pulling phase: shoulder in extension, elbow in flexion. The evolution of the shoulder and elbow $\Gamma_{\MVC}(\theta_s(t),\theta_e(t))$ for a complete push/pull operation is illustrated in Fig.~\ref{fig:drillingMVC}. 

Once again, a discontinuity is observed at the change between pushing and pulling phases. In this case, the discontinuity results because in our mathematical model only one $\Gamma_{MVC(t)}$ was needed to formulate push and pull (not to be confused with the individual muscle  $\Gamma_{\MVC}$, which is continuous). 
\subsection{Muscle fatigue}
The fatigue parameter $k$ is assumed constant for a given operator and at a given joint, for any performed motion. The following mean  values for $k$, from a previous study \cite{Ma2009a} were used in this framework: $k_{\shoulder} = 0.17$ and $k_{\elbow} = 0.24$~\cite{Ma2012}. This setting of $k$ was obtained by an identification process, and was based on anthropomorphic data, maximal torques in flexion and in extension, and body dynamics representation. 

\begin{figure}[!ht]
	\begin{center}
	\subfigure[pushing and pulling phases]{\includegraphics[width=0.47\textwidth]{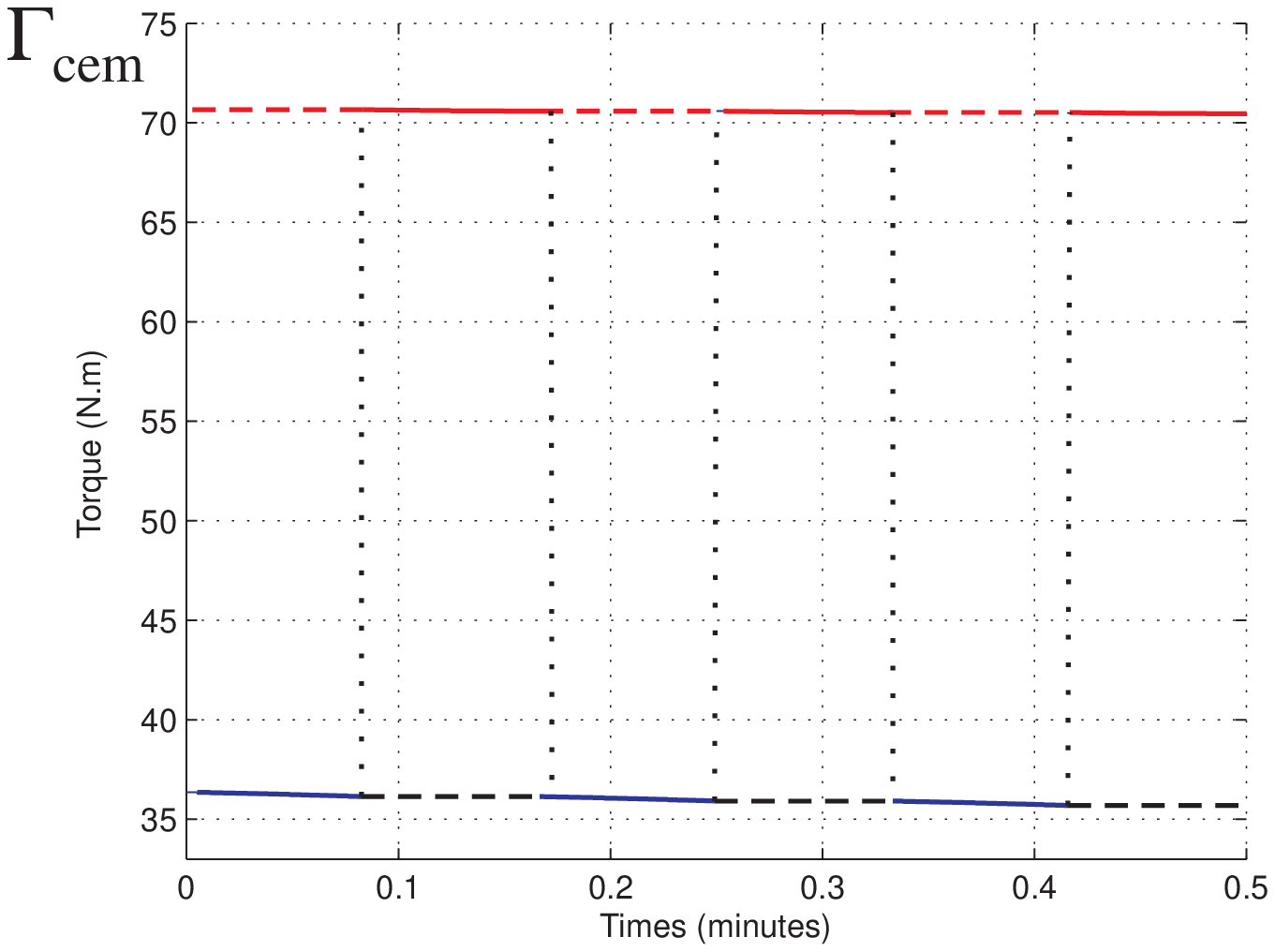}}\quad
	\subfigure[Zoom-in on pulling phases]{\includegraphics[width=0.47\textwidth]{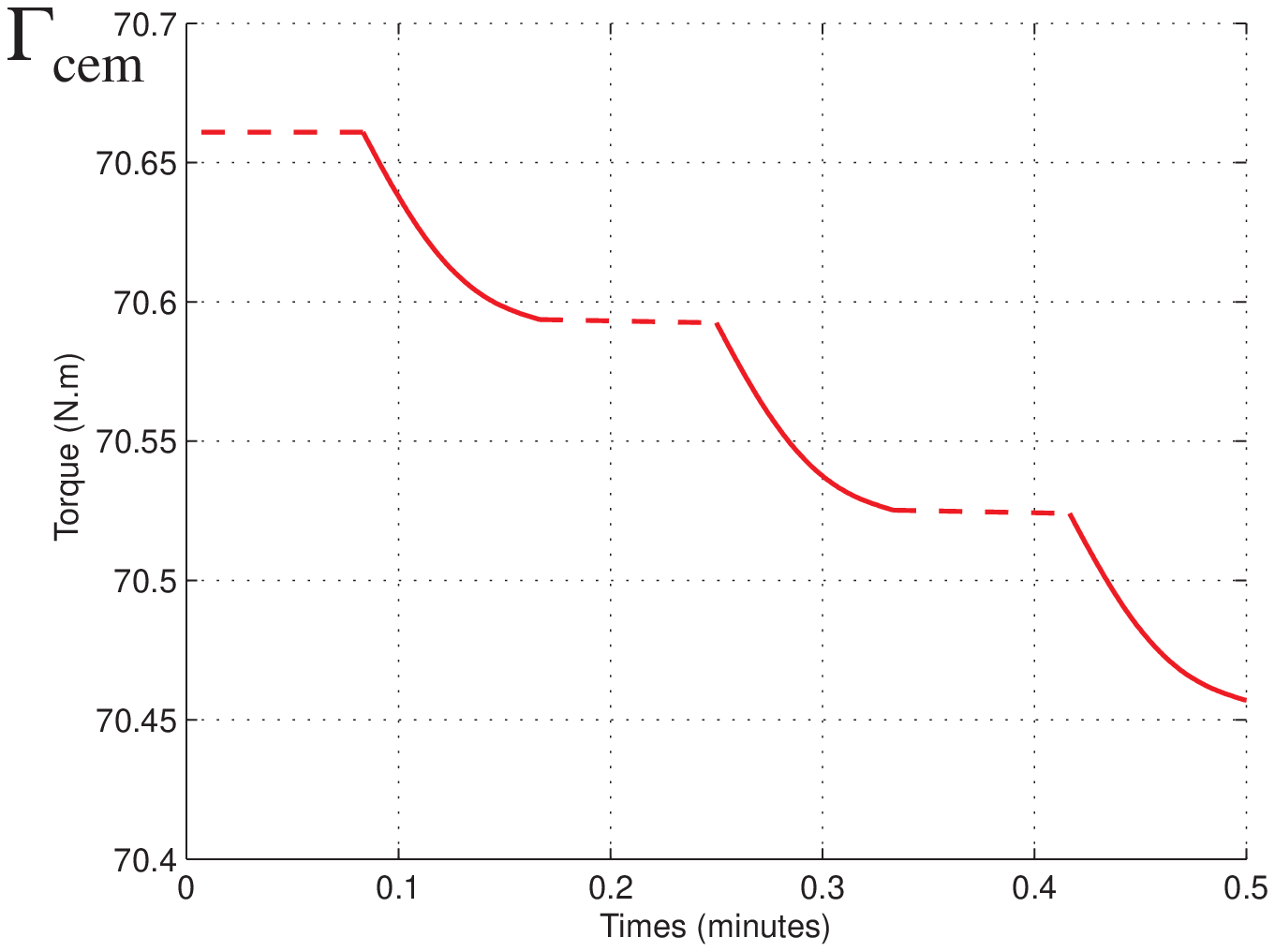}}\\
	\caption{Simulation of the decrease of muscle capacity $\Gamma_{\cem}$ at the elbow joint during repetitive push/pull operations for three cycles. The first phase is a pushing phase}
	\label{fig:drillingfatiguesimulationzoom}			 
	\end{center}
\end{figure}

Using the expressions of the shoulder and elbow joint torques extracted from Eq.~\eqref{eq:lagrangeclassic} and the matching joint capacities $\Gamma_{\MVC}(t)$, the fatigue model can be applied to these two joints for a complete push/pull cycle. Figure~\ref{fig:drillingfatiguesimulationzoom} illustrates the resulting fatigue predicted at the elbow joint for three push/pull cycles. As the external load remains low, the decrease in joint torque capacity $\Gamma_{\MVC}(t)$ is very slow and may not be easily observed in Fig.~\ref{fig:drillingfatiguesimulationzoom}(a), so a zoom-in on the pulling phases was realized and is shown in Fig.~\ref{fig:drillingfatiguesimulationzoom}(b). Here, the increase in fatigue can be seen. Similar results were obtained for the pushing phases. 

From Fig.~\ref{fig:drillingfatiguesimulationzoom}, we can see that the evolution of the fatigue (i.e., decrease in joint torque capacity $\Gamma_{\cem}$) is not continuous at the transition between pushing and pulling phases. The main reason for this discontinuity is that a single  $\Gamma_\MVC$ was formulated for push and pull muscle groups. 

The evolution of the fatigue within a pushing or a pulling phase mainly depends on the values taken by the torque $\Gamma_{\joint}$. The resulting behavior is then different for a pushing phase or for a pulling phase as $\Gamma_{\joint}$ is different according to the considered phase. 

\begin{figure}[!ht]
	\begin{center}
	\subfigure[Elbow]{\includegraphics[width=0.47\textwidth]{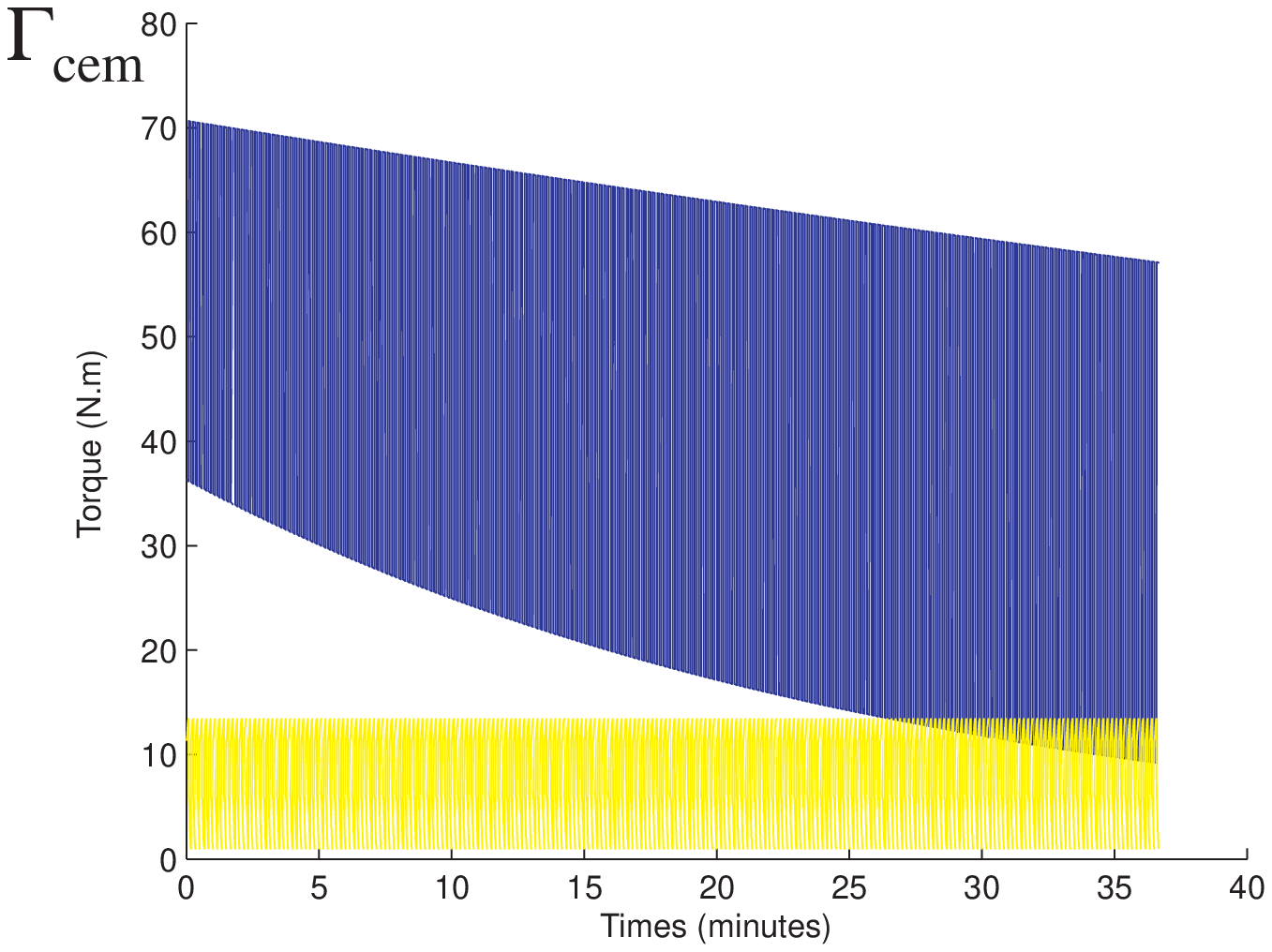}}\quad
	\subfigure[Shoulder]{\includegraphics[width=0.47\textwidth]{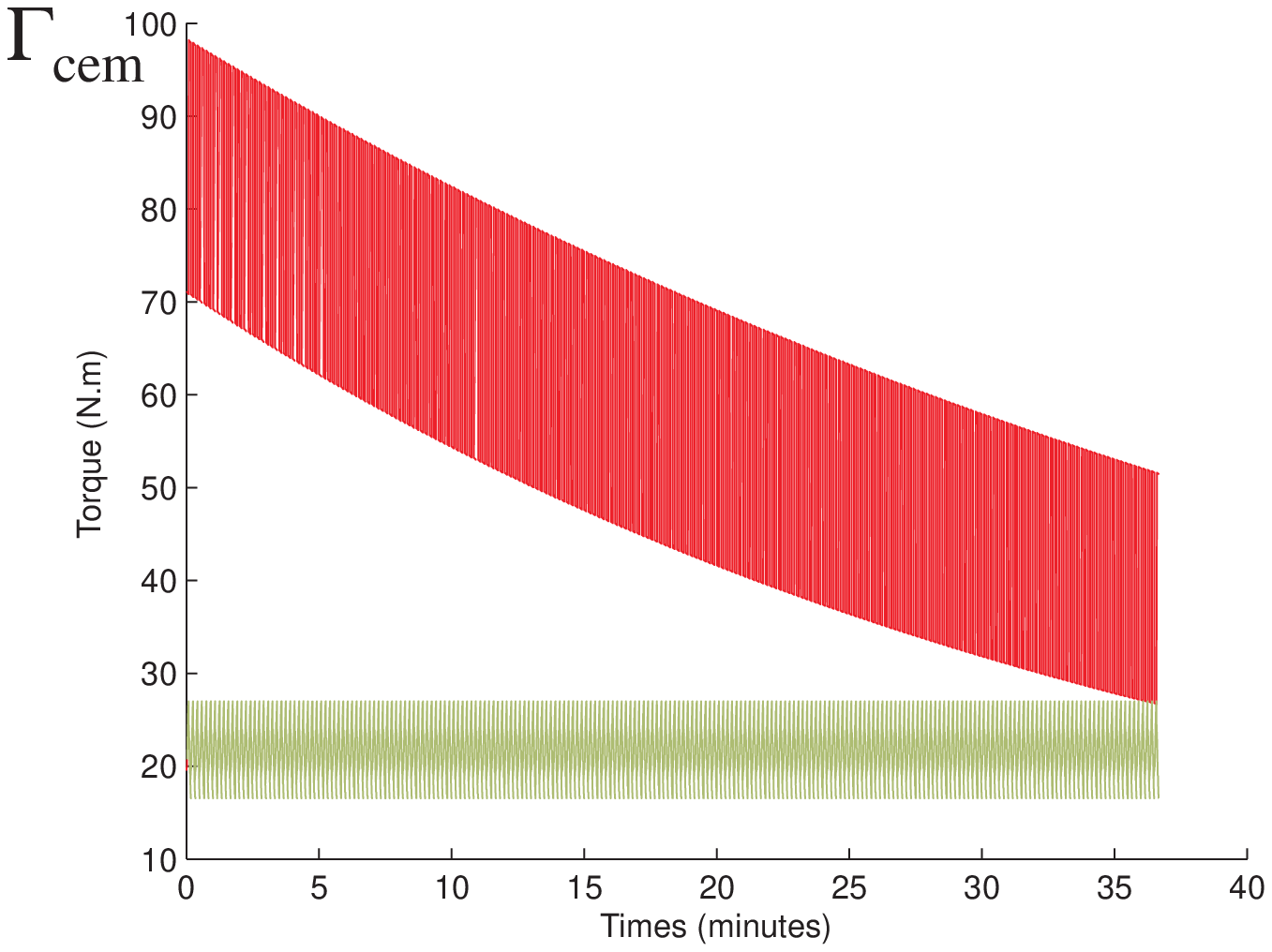}}
	\caption{Simulation of the decrease of muscle capacity at the shoulder and elbow joints during repetitive push/pull operations, and with the hand moving from $P_0 = [0.4, 0.1]$ to $P_f = [0.6, 0.1]$. The bottom curve represents the oscillations of the desired torque $\Gamma_{\joint}$ and the top curve represents the oscillations of the available torque $\Gamma_{\cem}$ with respect to the push and pull actions}
	\label{fig:drillingfatiguesimulation}
	\end{center}
\end{figure}

The effect of fatigue for such light loads should be evident after a long period of push/pull operations. The model was thus simulated for 37 minutes of exercise, and Figure~\ref{fig:drillingfatiguesimulation} illustrates the resulting fatigue predicted at the shoulder and elbow joints. For these two graphs, the bottom curve represents the oscillations of the desired torque $\Gamma_{\joint}$, which remains always the same, and the top curve represents the oscillations of the available torque $\Gamma_{\cem}$, which decreases with the increasing muscle fatigue. 
For both joints, the fatigue in pushing operation was more important than during the pulling operation. This is explained both by a less important joint capacity $\Gamma_{\cem}$ and a higher external load for the pushing phases. Using such graphs to predict potential MSD risks, such risk would seem most important when the two parts join (i.e. at around 26 minutes for the elbow joint and at around 35 minutes for the shoulder joint in this case). In this example too, MSD risk would seem higher at the elbow joint. Still, as the recovery model is not yet included in this approach, and thus in reality any risks would likely involve longer periods of activity. 
%

\subsection{Discussion}
\subsubsection{Effect of constant or variable joint capacity}
\begin{figure}[!ht]
	\centering
\subfigure[Elbow]{\includegraphics[width=0.47\textwidth]{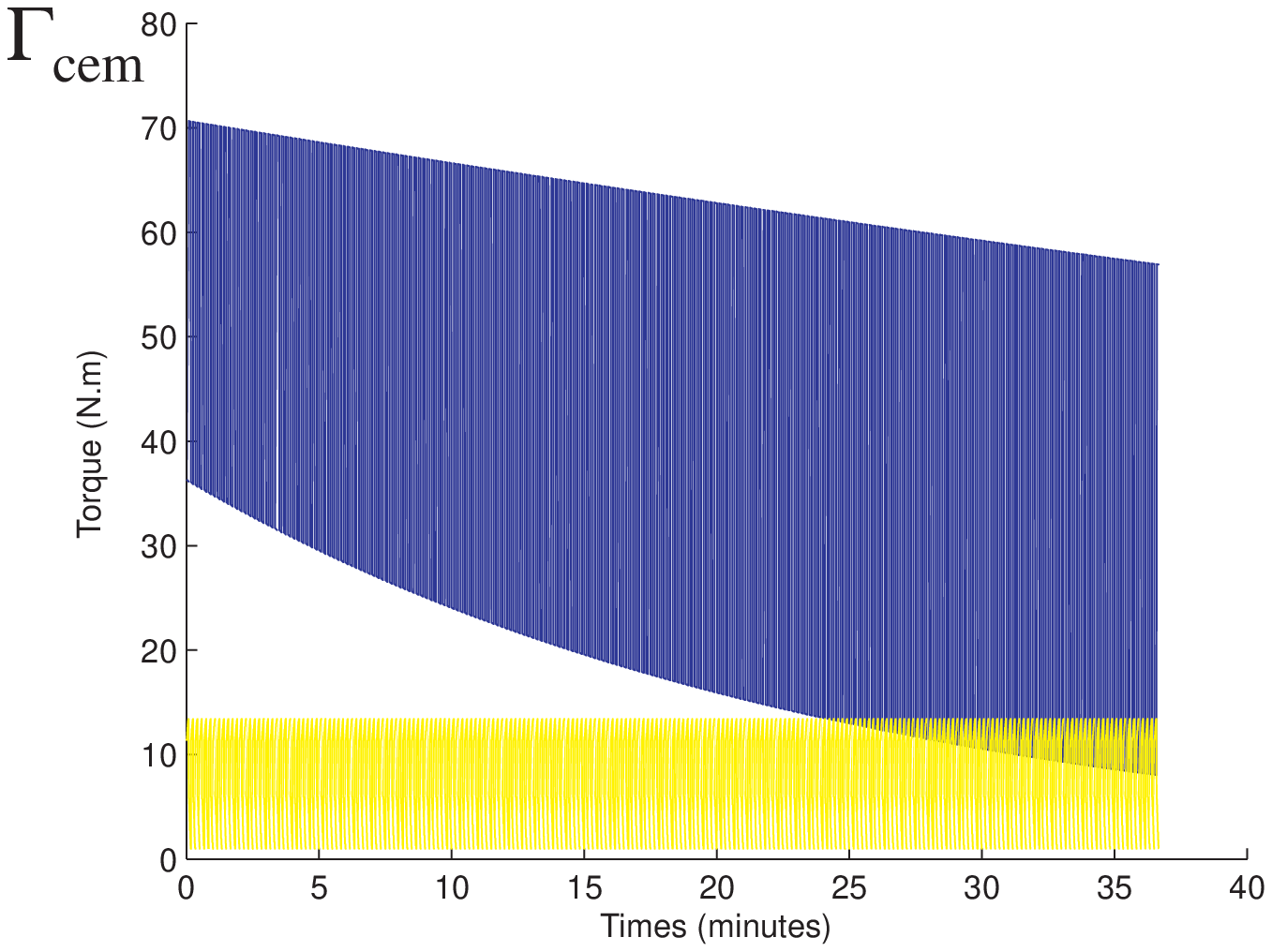}}\quad
\subfigure[Shoulder]{\includegraphics[width=0.47\textwidth]{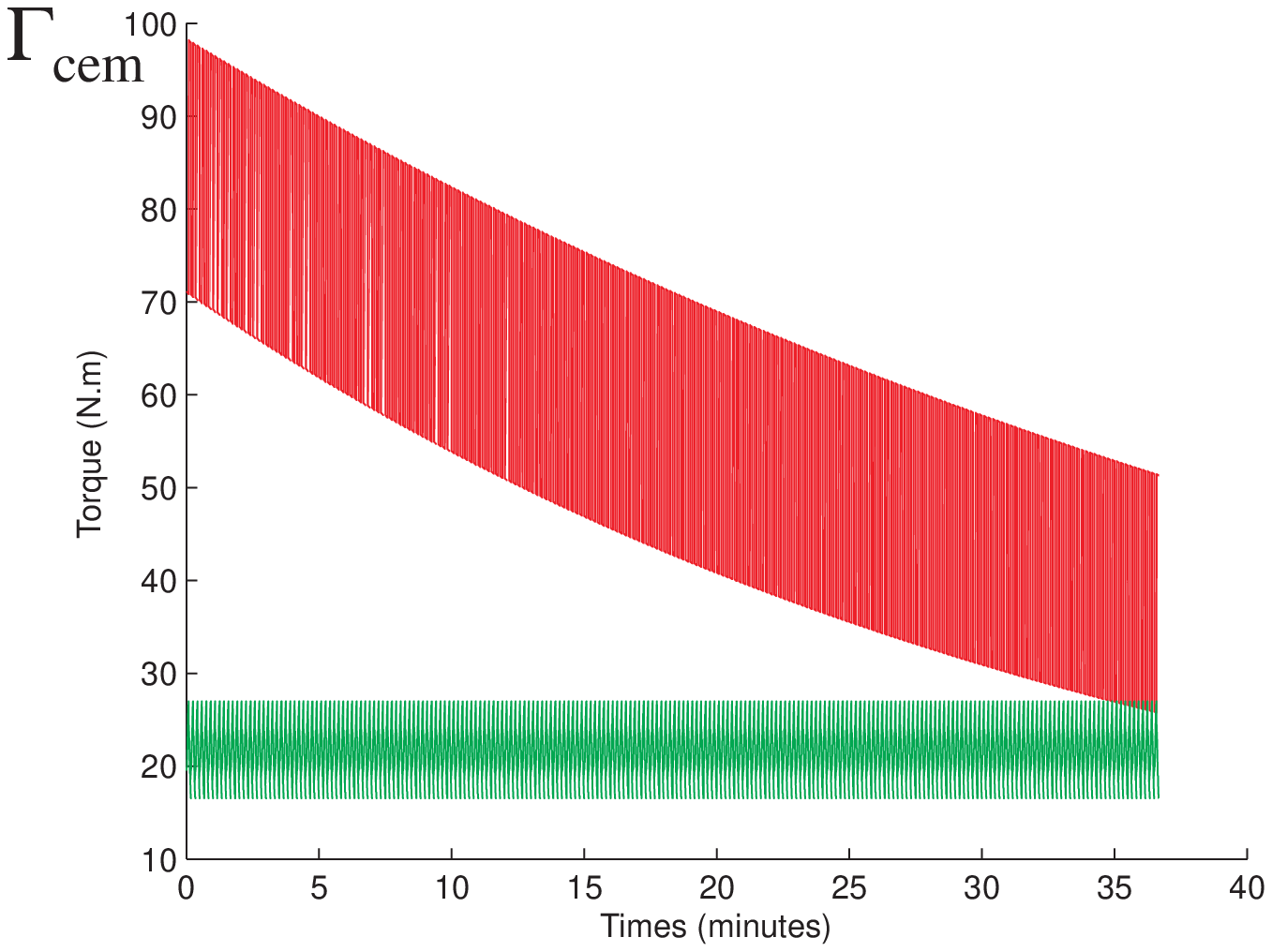}}
	\caption{Simulation of the decrease in muscular capacity of the elbow and shoulder when $\Gamma_{\MVC}$  is a constant. The bottom curve represents the oscillations of the desired torque $\Gamma_{\joint}$ and the top curve represents the oscillations of the available torque $\Gamma_{\cem}$ with respect to the push and pull actions}
	\label{fig:drillingfatiguesimulationMVCconstante}
\end{figure}

In this paper, the joint capacity $\Gamma_{\MVC}$ was introduced as a variable to build a model of fatigue in repetitive push/pull operations. To highlight the difference between the use of $\Gamma_{\MVC}$ as constant  or variable (i.e. depending on the posture), we compare the fatigue graphs obtained in the two situations. The minimum value of $\Gamma_{\MVC}$ obtained for all the simulated configurations was set as the reference constant value; this value will produce the greatest muscle fatigue. Figure~\ref{fig:drillingfatiguesimulationMVCconstante} shows the level of fatigue  in the situation when $\Gamma_{\MVC}$ is set as a constant.

The difference between the curves obtained by considering a variable value (figure~\ref{fig:drillingfatiguesimulation}) and a constant value (figure~\ref{fig:drillingfatiguesimulationMVCconstante}) is very small in this framework. The level of fatigue is slightly greater in the case when $\Gamma_{\MVC}$ is constant, because the worst situation was considered. An increased risk at the elbow appears after 23 minutes instead of 25 minutes, which is almost identical. So, it seems that the first approximation of a constant value for the joint capacity is adequate to evaluate the muscle fatigue, if including a slight allowance for the prediction of MSD. However, the current work setting may have an influence on this observation, and a general sensitivity analysis should be conducted to establish a more general conclusion that using a constant $\Gamma_{\MVC}$ is adequate to evaluate the muscle fatigue.
\subsubsection{Effect of two push/pull tasks}
\begin{figure}[!ht]
	\centering \subfigure[\label{fig:drillinglimiteelbowfatiguediscussion}Elbow]{\includegraphics[width=0.47\textwidth]{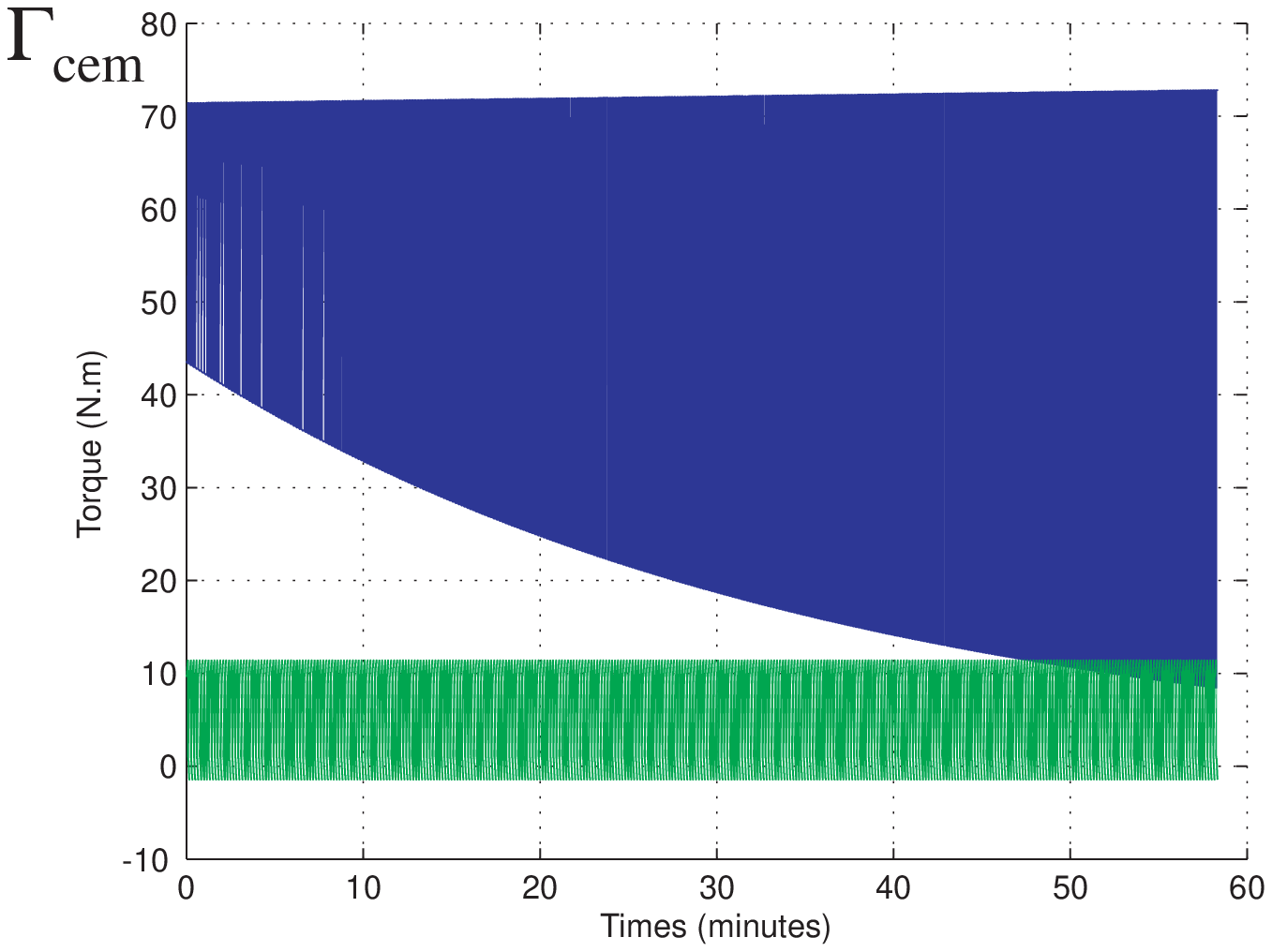}}\quad
\subfigure[Shoulder]{\includegraphics[width=0.47\textwidth]{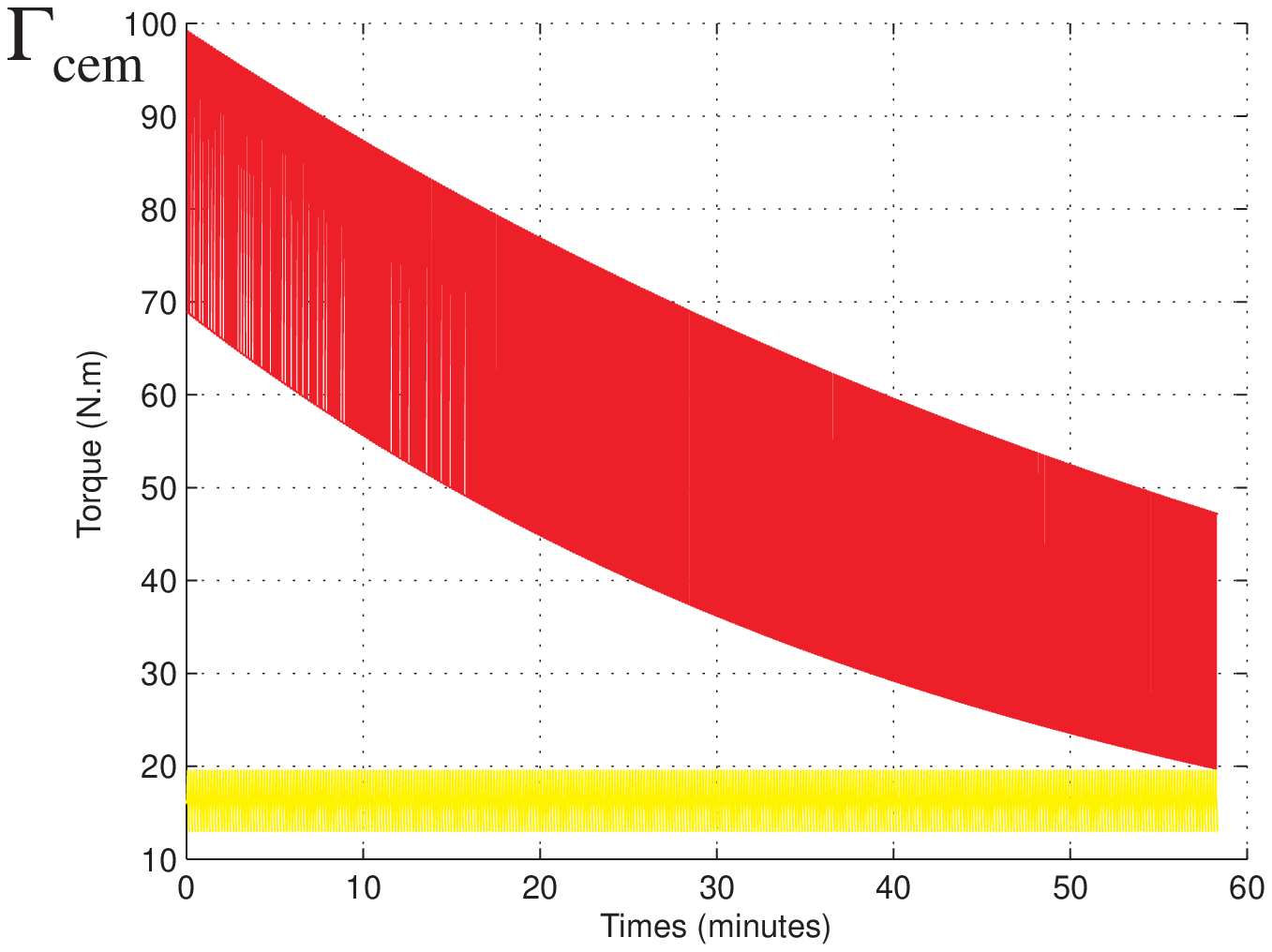}}
	\caption{Evolution of fatigue for the hand moving from $P_0=[0.3,0.1]$ to $P_f=[0.4,0.1]$. The bottom curve represents the oscillations of the desired torque $\Gamma_{\joint}$ and the top curve is the oscillations of the available torque $\Gamma_{\cem}$ with respect to the push and pull actions}
	\label{fig:drillinglimitefatiguediscussion}
\end{figure}

The hand movement amplitude was previously considered to be 20~[cm], moving from $P_0=[0.4,0.1]$ to $P_f=[0.6,0.1]$~[m]. If the initial and final positions are modified to $P_0=[0.3,0.1]$ and $P_f=[0.4,0.1]$~[m], as depicted in Fig.~\ref{fig:PushPullPostureDiscussion}, the resulting fatigue curves are shown in Fig.~\ref{fig:drillinglimitefatiguediscussion}.

\begin{figure}[!ht]
	\centering	
	\subfigure[$\Gamma_{\MVC}$]{\includegraphics[width=0.47\textwidth]{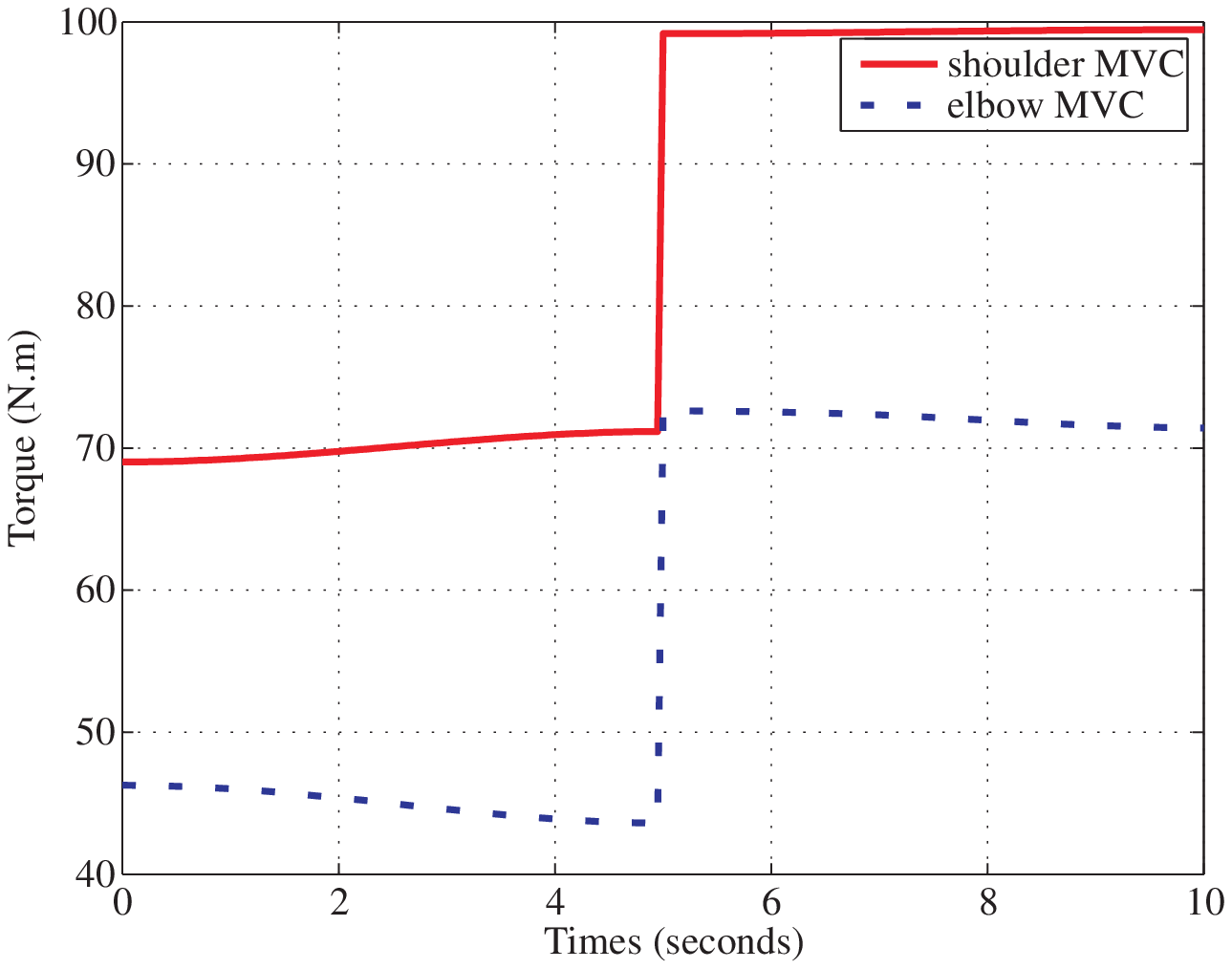}}\quad
	\subfigure[$\Gamma_{\joint}$]{\includegraphics[width=0.47\textwidth]{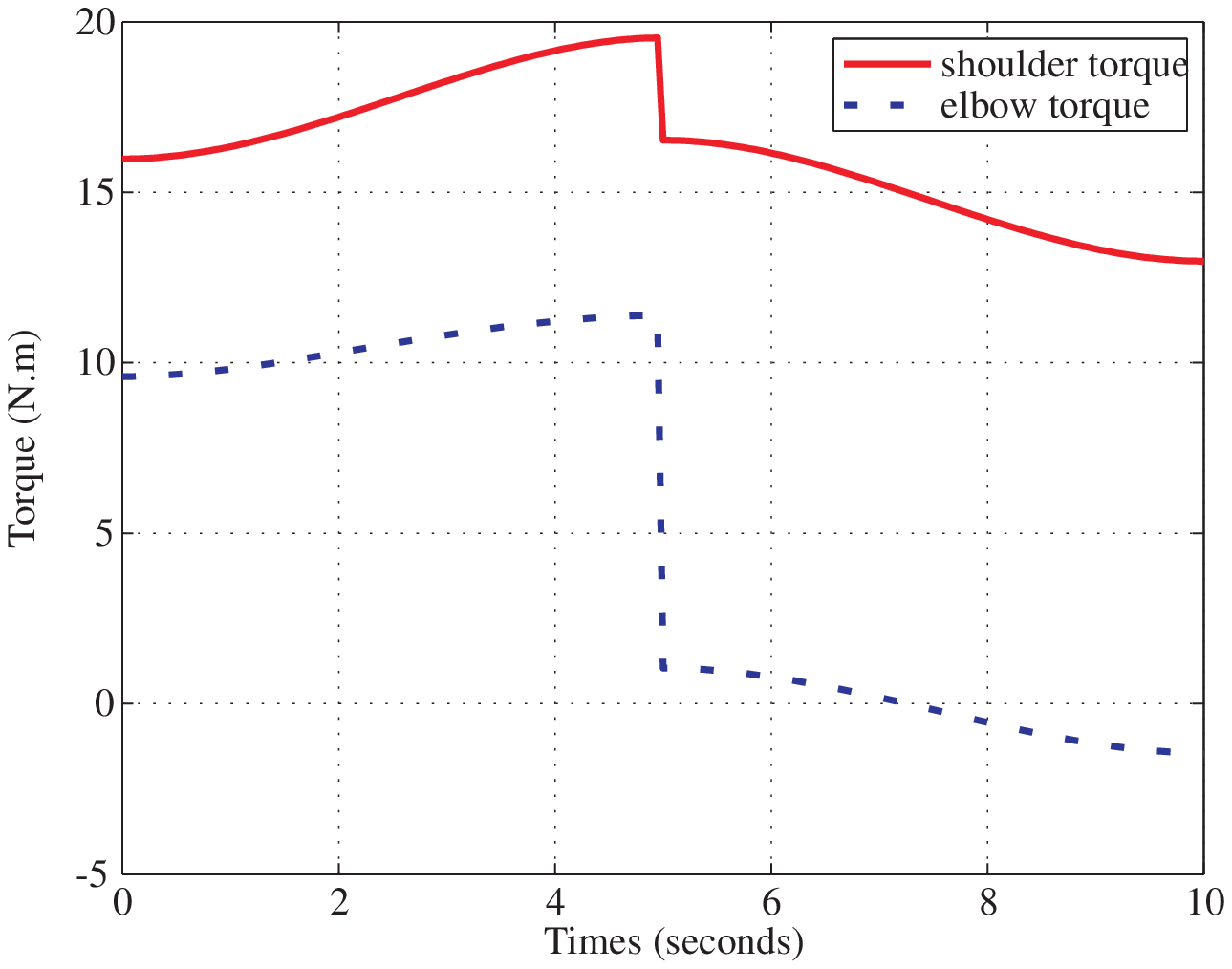}}\quad       	
	\caption{Evolution of the input torques at the elbow and shoulder for the hand moving from $P_0=[0.3,0.1]$ to $P_f=[0.4,0.1]$}
	\label{fig:drillinglimitetorquediscussion}
\end{figure}

As expected, the two push/pull task configurations generate different fatigue. The simulation was run this time for a sixty minutes of exercise. Still, Fig.~\ref{fig:drillinglimiteelbowfatiguediscussion} demonstrates that the fatigue curve for the elbow in the pulling phase is slightly increased: this may be observed on the upper values of the top oscillations (Fig.~\ref{fig:drillinglimiteelbowfatiguediscussion}), starting at 71 [m.s] and ending at 73 [m.s]. This may seem an unrealistic result,  because performing an operation should in theory reduce the joint capacity  $\Gamma_{\cem}$. To better understand the reason for this behavior, let us observe the torque evolution $\Gamma_{\joint}$ and joint capacity $\Gamma_{\MVC}$ for the studied operation, illustrated in Fig.~\ref{fig:drillinglimitetorquediscussion}. The second phase of Figure\ref{fig:drillinglimitetorquediscussion} is the pushing phase.  Figure\ref{fig:drillinglimitetorquediscussion}(b) demonstrates that the torque level is close to zero at the elbow joint, thus minimal fatigue is generated for the push muscle group at the elbow joint.
This particular finding may have workplace implications.  While the differences between the two push/pull task configurations used in the simulation are seemingly small, the end effect on fatigue can be dramatic (i.e. for the push muscle group at the elbow joint). Thus by configuring workplace tasks carefully, it may be possible to reduce the workload and/or fatigue for the most vulnerable muscles and reduce MSDs.
\section{Conclusions}
In this article, a quasi-static model of muscle fatigue was proposed as an extension of a previous model of static fatigue, by incorporating a variable joint capacity  $\Gamma_{\MVC}$ as a function of the operator's posture. The model was applied to simulate repetitive push/pull operations with light external loads that may be observed in an industrial framework, such as a classical drilling operation. From the simulation of fatigue during one task configuration, we found that fatigue of the elbow appears faster than the shoulder. This result is in agreement with the observation that most arm MSDs appear at the elbow joint. Indeed, the muscle strength of the elbow is lower than at the shoulder. However, for the second task configuration, we found minimal fatigue for the elbow. This last observation demonstrates that it is possible to reduce MSDs by optimizing workplace tasks. 

\bibliographystyle{plain}
\bibliography{Bibliography}
\end{document}